\pdfoutput=1
\documentclass{article} % For LaTeX2e
\usepackage{iclr2016_conference,times}
\usepackage[english]{babel}
\usepackage{microtype}
\usepackage{csquotes}
\usepackage[inline]{enumitem}
\usepackage{multirow}
\usepackage{booktabs}
\usepackage{caption,subcaption}
\usepackage{mathtools}
\usepackage{amsfonts}
\usepackage[hidelinks]{hyperref}

\iclrfinalcopy

\title{Variable Rate Image Compression with\\ Recurrent Neural Networks}

\author{
George Toderici, Sean M. O'Malley, Sung Jin Hwang, Damien Vincent\\
\texttt{\{gtoderici, smo, sjhwang, damienv\}@google.com}\\
\And David Minnen, Shumeet Baluja, Michele Covell \& Rahul Sukthankar\\
\texttt{\{dminnen, shumeet, covell, sukthankar\}@google.com}\\
\\
Google\\
Mountain View, CA, USA
}

\DeclareMathOperator{\sigm}{sigm}

\begin{document}

\renewcommand\sectionautorefname{Section}
\renewcommand\subsectionautorefname{Section}

\maketitle

\begin{abstract}
A large fraction of Internet traffic is now driven by requests from mobile devices with relatively small screens and often stringent bandwidth requirements.
Due to these factors, it has become the norm for modern graphics-heavy websites to transmit low-resolution, low-bytecount image previews (thumbnails) as
part of the initial page load process to improve apparent page responsiveness. Increasing thumbnail compression beyond
the capabilities of existing codecs is therefore a current research focus, as any byte savings will significantly enhance the experience of
mobile device users. Toward this end, we propose a general framework for variable-rate image compression and a novel
architecture based on convolutional and deconvolutional LSTM recurrent networks. Our models
address the main issues that have prevented autoencoder neural
networks from competing with existing image compression algorithms: (1) our
networks only need to be trained once (not per-image), regardless of
input image dimensions and the desired compression rate; (2) our networks
are progressive, meaning that the more bits are sent,
the more accurate the image reconstruction; and (3) the proposed
architecture is at least as efficient as a standard purpose-trained autoencoder for a
given number of bits. On a large-scale benchmark of 32$\times$32 thumbnails,
our LSTM-based approaches provide better visual quality than
(headerless) JPEG, JPEG2000 and WebP, with a storage size that is reduced by 10\% or
more.

\end{abstract}

\section{Introduction}

The task of image compression has been thoroughly examined over the years by
researchers and teams such as the Joint Pictures Experts Group, who designed the
ubiquitous JPEG and JPEG 2000 \citep{jpeg2000} image formats. More recently, the
WebP algorithm was proposed in order to further improve image compression
rates~\citep{webp:2015},
especially for the high-resolution images that have become more common in
recent years. All these efforts approach the compression problem from
an empirical standpoint: human experts design various heuristics to reduce the
amount of information needing to be retained, then determine ways to transform
the resulting data in a way that's amenable to lossless compression.
As this work is almost exclusively focused on the compression of large
images, low-resolution thumbnail images are usually ignored (and even harmed, e.g.,
by requiring more data in file headers).

Standard image compression algorithms tend to make assumptions about image
scale. For example, we usually assume that a patch from a high-resolution
natural image will contain a lot of redundant information. In fact, the higher-resolution
an image is, the more likely it is that its component patches will contain mostly
low-frequency information. This fact is exploited by most image codecs
and, as such, these codecs tend to be very efficient at compressing high-resolution
images. However, such assumptions are broken when creating thumbnails from high-resolution
natural images, as a patch taken from a thumbnail is much more likely
to contain difficult-to-compress, high-frequency information.

Large-scale compression of thumbnails (e.g., 32$\times$32 images) is an
important application, both in terms of reducing disk storage and making
better use of limited Internet bandwidth. Enormous numbers of thumbnails are
currently transmitted across the web for page previews, photo galleries, search engine results,
and numerous other applications. As such, any improvements to thumbnail compression
will significantly improve the experience of users accessing content over
low-bandwidth connections.

In recent years, neural networks have become commonplace to perform tasks that
had for decades been accomplished by {\em ad hoc} algorithms and heuristics. For
instance, in image recognition and object detection, the current
state-of-the-art algorithms are all based on neural networks. It is only natural
to ask if we can also employ this powerful class of methods to further improve
the task of image compression, especially for image sizes for which we
do not have carefully designed, hand-tuned compressors.

If we consider an image codec broadly as an analysis/synthesis problem with
a bottleneck in the middle, then we can find a significant body of research aimed toward
teaching neural networks to discover compressive representations. Most of this
work \citep[e.g.,][]{denton2015,gregor2015}, has been on synthesis of
small images: often 32$\times$32 in part due to
CIFAR10 \citep{Krizhevsky09learningmultiple}.
Much of this work has focused on a class of neural networks known
as {\em autoencoders} \citep{Krizhevsky2011}. However, standard autoencoders
operate under a number of hard constraints that have so far made them
infeasible as a drop-in replacement for standard image codecs. Some of these
constraints are that variable-rate encoding is typically not possible (one
network is trained per compression rate); the visual quality of the output
is hard to ensure; and they're typically trained for a particular scale, being
able to capture redundancy only at that scale.

We explore several different ways in which neural network-driven image
compression can improve compression rates while allowing similar flexibility
to modern codecs. To achieve this flexibility, the network
architectures we discuss must meet all of the following requirements:
\begin{enumerate*}[label=(\arabic*)]
\item the compression rate should be capable of being restricted to a prior bit
budget;
\item the compressor should be able to encode simpler patches more cheaply (analogously to modern codecs which may
allocate more bits to areas of the image which contain important visual features); and
\item the model should be able to learn from large amounts of existing imagery
in order to optimize this compression process toward real-world data.
\end{enumerate*}

\section{Related Work}

The basic principles of using feed-forward neural networks for image compression
have been known for some time \citep{Jiang1999}. In this context, networks can
assist or even entirely take over many of the processes used as part of a
traditional image compression pipeline: to learn more efficient frequency
transforms, more effective quantization techniques, improved predictive coding, etc.

More recently, autoencoder architectures~\citep{Hinton2006} have become
viable as a means of implementing end-to-end compression. A typical compressing
autoencoder has three parts:
\begin{enumerate*}[label=(\arabic*)]
\item an {\em encoder} which consumes an input (e.g., a fixed-dimension image or
patch) and transforms it into
\item a {\em bottleneck} representing the compressed data, which can then be
transformed by
\item a {\em decoder} into something resembling the original input.
\end{enumerate*}
These three elements are trained end-to-end, but during deployment the encoder
and decoder are normally used independently.

The bottleneck is often simply a flat neural net layer, which allows the
compression rate and visual fidelity of the encoded images to be controlled by
adjusting the number of nodes in this layer before training. For some types of
autoencoder, encoding the bottleneck as a simple bit vector can be
beneficial~\citep{Krizhevsky2011}. In neural net-based classification tasks,
images are repeatedly downsampled through convolution and pooling operations,
and the entire output of the network might be contained in just a single node.
In the decoder half of an autoencoder, however, the network must proceed in
the opposite direction and convert a short bit vector into a much larger image
or image patch. When this upsampling process is spatially-aware, resembling a
\enquote{backward convolution,} it is commonly referred to as {\em
deconvolution}~\citep{Long2014}.

Long short-term memory (LSTM) networks are a type of recurrent neural
network~\citep{lstm:1997} that have proven very successful for tasks such as
speech recognition~\citep{Graves2013} and machine
translation~\citep{Sutskever2014}. Many extensions to the standard LSTM model
are possible including explicitly incorporating spatial information, which
leads to various types of convolutional LSTMs~\citep{Shi2015}
that may be better suited for image compression. We experiment with such
models and also try simpler recurrent architectures that use the residual
error of one autoencoder as the input to another.

\section{Variable Rate Compression Architectures}

We start by describing a general neural network-based compression
framework and then discuss the details of multiple instantiations of
this architecture. Each subsection describes a different architecture
that builds on the previous model and improves the compression
results.

For each architecture, we will discuss a function $E$ that takes an image
patch as input and produces an encoded representation. This representation is
then processed by a binarization function $B$, which is the same across
architectures, and is discussed in \autoref{sec:bin}. Finally, for each
architecture we also consider a decoder function $D$, which takes the binary
representation produced by $B$ and generates a reconstructed output
patch. Taken together, these three components form an autoencoder,
$x'=D(B(E(x)))$, which is the basic building block for all of the compression
networks.

For all architectures, an offset and scale are applied to the 8-bit RGB input
images to give a range of values between -0.9 and 0.9. This range is compatible
with the values that can be emitted by $\tanh$.

\subsection{Image Compression Framework}
The neural network architectures that we use share the same conceptual stages:
an encoder network, followed by a quantizer, and a decoder network. In addition,
our framework is tuned for image compression and supports variable
compression rates without the need for retraining or for storing multiple
encodings of the same image.

To make it possible to transmit incremental information, the design should
take into account the fact that image decoding will be progressive. With this
design goal in mind, we can consider architectures that are built on top of
residuals with the goal of minimizing the residual error in the reconstruction
as additional information becomes available to the decoder.

Formally, we chain multiple copies of a residual autoencoder, $F_t$, defined
as:
\begin{equation}
 F_t(r_{t-1}) = D_{t}(B(E_{t}(r_{t-1}))).
\end{equation}

This chaining is explicit, in the case of our feed-forward-only
networks (\autoref{sec:elmer} and \autoref{sub:dewey}) and is
implicit, through the recurrent structure, in the case of our LSTM
networks (described in
\autoref{sub:woody} and \autoref{sub:donalduck}).
In all cases, we set $r_0$ to be equal to the original input patch, and then
$r_{t}$ for $t>0$ represents the residual error after $t$ stages. For non-LSTM
architectures (described in Sections~\ref{sec:elmer} and~\ref{sub:dewey}),
$F_t$ has no memory, and so we only expect it to predict the residual
itself. In this case, the full reconstruction is recovered by summing over all
of the residuals, and each stage is penalized due to the difference between
the prediction and the previous residual:
\begin{equation}
 r_{t} = F_t(r_{t-1}) - r_{t-1}.
\end{equation}

On the other hand, LSTM-based architectures (described in
Sections~\ref{sub:woody} and~\ref{sub:donalduck}) do hold state, and so we
expect them to predict the original image patch in each stage. Accordingly, we
compute the residual relative to the original patch:
\begin{equation}
 r_{t} = F_t(r_{t-1}) - r_0.
\end{equation}

In both cases, the full, multi-stage network is trained by minimizing $\lVert
r_{t}\rVert_2^2$ for $t = 1 \ldots N$, where $N$ is the total number of residual
autoencoders in the model.

%% Formally, we set $r_{t} = D_{t}(B(E_{t}(r_{t-1}))) - p_{t}$ and minimize
%% $\lVert r_{t}\rVert_2$. The definition of $p_{t}$ depends on the type of
%% model:
%% %
%% \begin{equation}
%%   p_{t} =
%%   \begin{cases}
%%     r_{t-1} & \text{for non-LSTM models},\\
%%     r_{0} & \text{for LSTM-based models}.
%%   \end{cases}
%% \end{equation}

%% In all cases, we set $r_0$ to be equal to the original input patch, and then
%% $r_{t}$ represents the residual error after $t$ stages. The
%% $D_{t}(B(E_{t}(r_{t-1})))$ structure represents a residual autoencoder, and we
%% can chain multiple copies to learn a single model capable of generating a
%% variable rate encoding of image patches.

\subsection{Binary Representation}
\label{sec:bin}

In our networks, we employ a binarization technique first proposed
by~\cite{williams1992simple}, and similar to~\cite{Krizhevsky2011} and
\cite{binarization}.  This binarization has three benefits:
\begin{enumerate*}[label=(\arabic*)]
\item bit vectors are trivially serializable/deserializable for image
  transmission over the wire,
\item control of the network compression rate is achieved simply by putting
  constraints on the bit allowance, and
\item a binary bottleneck helps force the network to learn efficient
  representations compared to standard floating-point layers, which may have
  many redundant bit patterns that have no effect on the output.
\end{enumerate*}

The binarization process consists of two parts. The first part consists of
generating the required number of outputs (equal to the desired number of
output bits) in the continuous interval $[-1, 1]$. The second part involves
taking this real-valued representation as input and producing a discrete
output in the set $\{-1, 1\}$ for each value.

For the first step in the binarization process, we use a fully-connected layer
with $\tanh$ activations.  For the second part, following \citet{raiko:2015},
one possible binarization $b(x)$ of $x \in [-1,1]$ is defined as:
\begin{equation}
  b(x) = x + \epsilon \quad \in \{-1, 1\},
\end{equation}
\begin{equation}
  \epsilon \sim
  \begin{cases}
    1 - x & \text{with probability $\frac{1 + x}{2}$},\\
    - x - 1 & \text{with probability $\frac{1 - x}{2}$},
  \end{cases}
\end{equation}
where $\epsilon$ corresponds to quantization noise.  We will use the
regularization provided by the randomized
quantization to allow us to cleanly backpropagate
gradients through this binarization layer.

Therefore, the full binary encoder function is:
\begin{equation}
  B\left(x\right) = b\left(\tanh(W^{\text{bin}}x + b^{\text{bin}})\right).
\end{equation}

where $W^{\text{bin}}$ and $b^{\text{bin}}$ are the standard linear weights
and bias that transform the activations from the previous layer in the
network.
In all of our models, we use the above formulation for the forward
pass. For the backward pass of back-propagation, we take the derivative
of the expectation \citep{raiko:2015}. Since $\mathbb E[b(x)] = x$
for all $x \in [-1, 1]$, we pass the gradients through $b$ unchanged.

In order to have a fixed representation for a particular input, once the
networks are trained, only the most likely outcome of $b(x)$ is considered
and $b$ can be replaced by $b^{\text{inf}}$ defined as:
\begin{equation}
  b^{\text{inf}}\left(x\right) =
  \begin{cases}
    -1 & \text{if $x < 0$,}\\
    +1 & \text{otherwise.}
  \end{cases}
\end{equation}

The compression rate is determined by the number of bits generated in each
stage, which corresponds to the number of rows in the $W^{\text{bin}}$ matrix,
and by the number of stages, controlled by the number of repetitions of the
residual autoencoder structure.

\subsection{Feed-Forward Fully-Connected Residual Encoder}
\label{sec:elmer}

In the simplest instantiation of our variable rate compression architecture,
we set $E$ and $D$ to be composed of stacked fully-connected layers. In order
to make the search for architectures more feasible we decided to set the
number of outputs in each fully-connected layer to be constant (512) and only
used the $\tanh$ nonlinearity.

\begin{figure*}[tbp]
  \centering
  \includegraphics[width=0.98\linewidth]{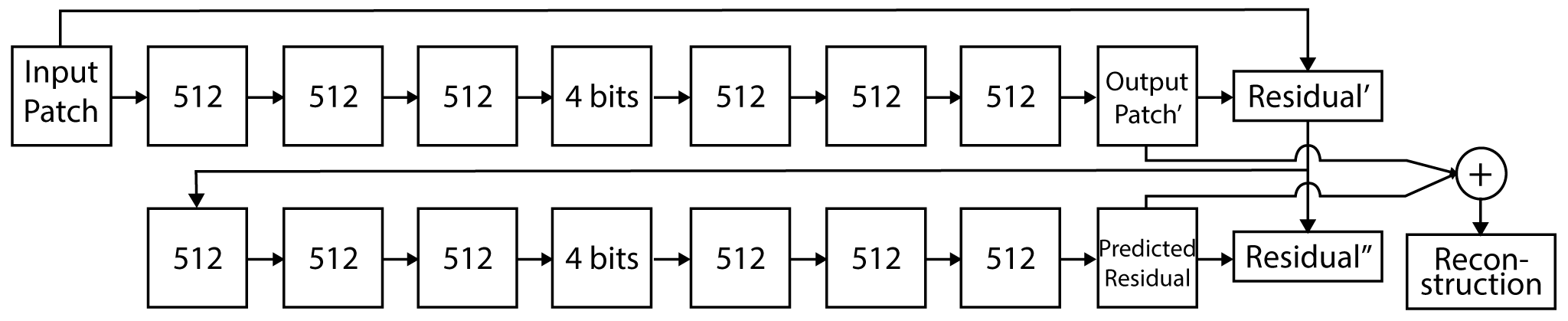}
  \caption{The fully-connected residual autoencoder. We depict
  a two-iteration architecture, with the goal of the first iteration being to
  encode the original input patch and the goal of the second iteration
  being to encode the residual from the first level's reconstruction.
  In our 64-bit results, reported in \autoref{tab:results32}, we
  have 16 iterations giving 4 bits each.
 The blocks marked with 512 are fully-connected neural network layers with
  512 units and $\tanh$ nonlinearities. The loss applied to the
  residuals in training is a simple L2 measure.}
  \label{fig:elmer}
\end{figure*}

Given that $E$ and $D$ can be functions of the encoding stage number, and
since the statistics of the residuals change when going from stage $t$ to
$t+1$ we considered two distinct approaches: in the first we share weights
across all stages, while in the second, we learn the distinct weights independently in
each stage.
The details of this architecture are given in \autoref{fig:elmer}.

\begin{figure*}[tbp]
  \centering
  \includegraphics[width=0.98\linewidth]{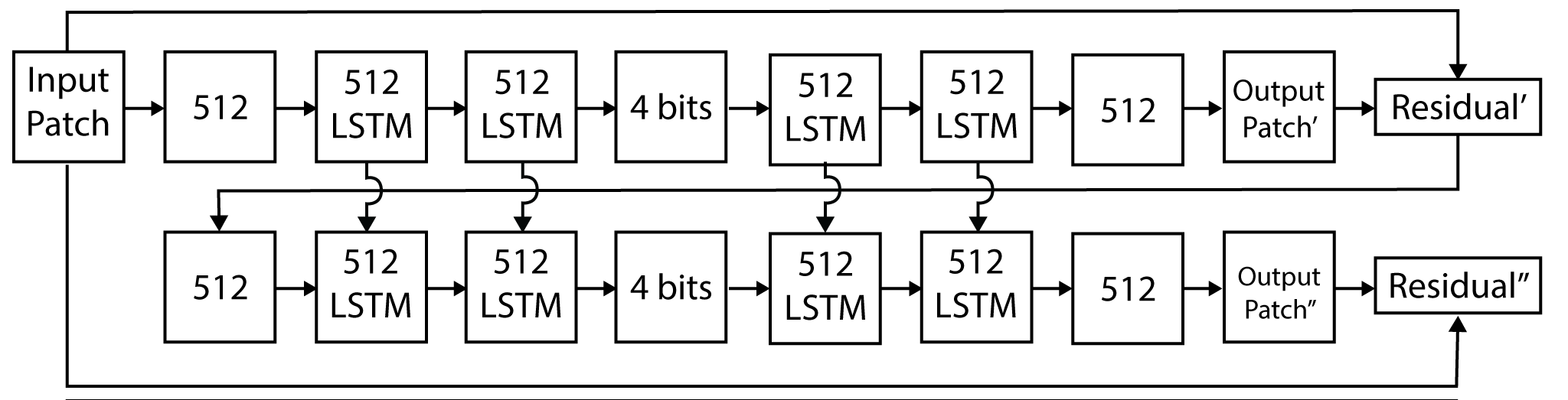}
  \caption{The fully-connected LSTM residual encoder. The 512 LSTM blocks
  represent LSTM layers with 512 units. This figure shows an unrolling
  of the LSTM, needed for training, to two time steps.  The actual
  architecture would have only the first row of blocks, with the
  functionality of the second row (and subsequent recursions) being
  realized by feeding the
  residual from the previous pass back into the first LSTM block.  For
  the results reported in Table~\ref{tab:results32}, this repeated
  feeding back was done 16 times, to generate 64 bit representations.
  The vertical connections between the LSTM stages in the unrolling
  shows the effect of the persistent memory instead each LSTM.
  The loss is applied to the residuals in training is a simple L2 measure.
  Note that in contrast to \autoref{fig:elmer}, in which the network after the first step is used
  to predict the previous step's residual error, in this LSTM architecture, each step
  predicts the actual output.}
  \label{fig:woody}
\end{figure*}

\begin{figure*}[tbp]
  \centering
  \includegraphics[width=0.98\linewidth]{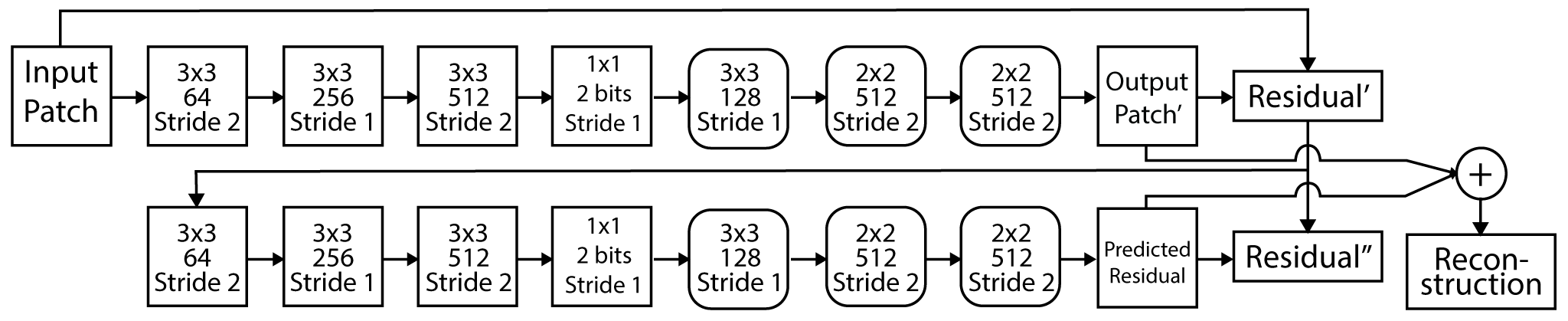}
  \caption{The convolutional / deconvolutional residual encoder. The
    convolutional layers are depicted as sharp rectangles, while the
    deconvolutional layers are depicted as rounded rectangles. The loss is
    applied to the residuals.}
  \label{fig:dewey}
\end{figure*}

\subsection{LSTM-based Compression}
\label{sub:woody}

In this architecture, we explore the use of LSTM models for both the encoder
and the decoder. In particular, both $E$ and $D$ consist of stacked LSTM
layers.

Following the LSTM formulation and notation proposed
by~\cite{zaremba2014recurrent}, we use superscripts to indicate the layer
number, and subscripts to indicate time steps.  Let $h^l_t \in \mathbb R^n$
denote the hidden state of $l$-th LSTM layer at time step $t$.  We define
$T^{l}_{n} \colon \mathbb{R}^m \to \mathbb{R}^n$ to be an affine transform
$T^{l}_{n}(x) = W^{l}x + b^{l}$. Finally, let $\odot$ denote element-wise
multiplication, and let $h^{0}_{t}$ be the input to the first LSTM layer at
time step $t$.

Using this notation, the LSTM architecture can be written succinctly as
proposed by \cite{graves2013generating}:
\begin{equation}
\begin{pmatrix}
i \\
f \\
o \\
g
\end{pmatrix} = \begin{pmatrix}
\sigm \\
\sigm \\
\sigm \\
\tanh
\end{pmatrix} T^{l}_{4n} \begin{pmatrix}
h^{l-1}_t \\
h^l_{t-1}
\end{pmatrix},
\label{eqn:full}
\end{equation}
\begin{align}
  c^{l}_{t} & = f \odot c_{t-1}^{l} + i \odot g, \label{eqn:c} \\
  h_{t}^{l} & = o \odot \tanh\left(c_{t}^{l}\right), \label{eqn:h}
\end{align}
where $\sigm(x) = (1+\exp(-x))^{-1}$ denotes the sigmoid function.

In these equations, $\sigm$ and $\tanh$ are applied element-wise. This
alternate formulation of LSTM is useful because it reduces the numbers of
separate operations needed to evaluate one step, which allows for an efficient
implementation on GPU.

For the encoder, we use one fully-connected layer followed by two stacked LSTM
layers. The decoder has the opposite structure: two stacked LSTM layers
followed by a fully-connected layer with a $\tanh$ nonlinearity that predicts
RGB values (we omit this layer in the diagrams to reduce clutter).
The exact architecture used in the experiments is given
in \autoref{fig:woody} (minus the RGB conversion).

\subsection{Feed-Forward Convolutional/Deconvolutional Residual Encoder}
\label{sub:dewey}

\autoref{sec:elmer} proposed a fully-connected residual autoencoder. We
extend this architecture by replacing the fully-connected layers with
convolution operators in the encoder $E$ and with deconvolutional operators in
the decoder $D$.
The final layer of the decoder
consists of a 1$\times$1 convolution with three filters that converts the
decoded representation into RGB values. We depict this architecture in
Figure~\ref{fig:dewey} (minus the RGB conversion).

The deconvolutional operator is defined as the transpose of the convolutional
operator. Let $\otimes_k$ denote the convolutional operator with stride $k$,
and let $S_k$ denote the stride operator with stride factor $k$, i.e.,
$S_k(x)(i,j) = x(k \times i, k \times j)$ for 2D multi-channel image $x$ and pixel coordinate
$(i,j)$. Then $W \otimes_k x = S_k(W \otimes_1 x)$. Note that the transpose of
$S_k$ is the \enquote{inflation} operator $T_k$:
\begin{equation}
  T_k (x)(i,j) =
  \begin{cases}
    x(i/k, j/k) & \text{if $i,j$ are multiples of $k$,}\\
    0 & \text{otherwise.}
  \end{cases}
\end{equation}
Thus we can define the deconvolutional operator $\oslash_k$ with stride $k$ as
follows:
\begin{equation}
  W \oslash_k x = W \otimes_1 (T_k (x)).
\end{equation}

\subsection{Convolutional/Deconvolutional LSTM Compression}
\label{sub:donalduck}

The final architecture combines the convolutional and deconvolutional
operators with LSTM. We define convolutional LSTM by replacing the
transformation $T^l_{4n}$ in equation~\eqref{eqn:full} with convolutions plus
bias. Then the transformation function for convolutional LSTM with stride $k$
is
\begin{equation}
  T^{l}_{4n}\left(h^{l-1}_{t}, h^{l}_{t-1} \right)
  =  W^{l}_{1} \otimes_k h^{l-1}_{t}  + W^{l}_{2} \otimes_1 h^{l}_{t-1} + b^{l}.
\end{equation}
The subscript belonging to $T$ now refers to the depth (number of features) in
the output feature maps. Note that the second convolution term represents the
recurrent relation of convolutional LSTM so both its input and output must
have the same size. Therefore, when a convolutional LSTM has a stride greater
than one, the stride is only applied to the first convolution term, while the
second term is always computed with a stride of one. Finally, to build the
encoder for this architecture, we replace the second and third convolutional
layers from ~\autoref{fig:dewey} with convolutional LSTM layers.

For the decoder, we cannot replace all convolutional operations with
deconvolution due to the fact that the input to deconvolution often has a
different spatial dimension than the output. For the purposes of defining a
deconvolutional LSTM, $T_{4n}$ becomes
\begin{equation}
  T^{l}_{4n}\left(h^{l-1}_{t}, h^{l}_{t-1} \right)
  =  W^{l}_{d} \oslash_k h^{l-1}_{t}  + W^{l}_{c} \otimes_1 h^{l}_{t-1} + b^{l}.
\end{equation}
Here we use the subscripts $c$ and $d$ to differentiate between the weights
associated to the convolution and deconvolution operations. To construct the
deconvolutional LSTM decoder, we replace the second and third deconvolutional
layers of the deconvolutional decoder from \autoref{fig:dewey} with deconvolutional LSTM.

\subsection{Dynamic Bit Assignment}

For the non-convolutional approaches presented here, it is natural to assign a
varying number of bits per patch by allowing a varying number of iterations of
the encoder. This could be determined by a target quality metric (e.g., PSNR).
While not as natural, in the case of the convolutional approaches, a similar
method may also be employed. The input image needs to be split into patches, and
each patch processed independently, thereby allowing a different number of bits
per region. However, this approach has disadvantages that will be discussed at
the end of this paper.

\section{Experiments \& Analysis}

\subsection{Training}

In order to train the various neural network configurations, we used the Adam
algorithm proposed by~\cite{kingma:2014}. We experimented with learning rates
of $\{0.1, 0.3, 0.5, 0.8, 1\}$. The $L_2$ loss was normalized by the number of
pixels in the patch and also by the number of total time steps (i.e., number
of iterations unrolled) needed to fully encode the patch.
We employed no perceptual weighting to improve the compression for
evaluation under the SSIM measure. During training we used the unmodified $L_2$
error measure.

We experimented with the number of steps needed to encode each patch, varying
this from 8 to 16. For the fully connected networks, we chose to use 8 bits per
step for an 8$\times$8 patch, allowing us to fine tune the compression rate in
increments of 8 bits. When scaled up to a 32$\times$32 patch size, this allowed
us to control the compression in increments of 128 bits.

For the convolutional/deconvolutional networks, the encoders reduce the 32$\times$32
input patch down to 8$\times$8 through convolution operations with strides.
We experimented with a binary output of 2 bits per pixel at this resolution, yielding a tunable compression rate
with increments of 16 bytes per 32$\times$32 block.

\subsection{Evaluation Protocol and Metrics}

Evaluating image compression algorithms is a non-trivial task. The metric
commonly used in this context is the peak signal-to-noise ratio (PSNR), however,
PSNR is biased toward algorithms which have been tuned to minimize $L_2$ loss.
This would not be a fair comparison against methods like JPEG which have been
tuned to minimize a form of perceptual loss.

In our evaluation protocol we instead employ the Structural Similarity Index
(SSIM), a popular perceptual similarity measure proposed by~\cite{ssim}.
Since we're evaluating compression performance on small 32$\times$32 images, we
do not smooth the images (a typical preprocess for SSIM).
In addition, since we're interested in quantifying how well local details are
preserved, we split the images into 8$\times$8 patches and compute the SSIM on
each patch and on each color channel independently. The final score is the
average SSIM over all patches and channels.

\begingroup
\renewcommand{\arraystretch}{1.25}
\begin{table*}[tbp]
  \small
\newcommand\specialcellc[2][c]{\begin{tabular}[#1]{@{}c@{}}#2\end{tabular}}
\newcommand\specialcelll[2][c]{\begin{tabular}[#1]{@{}l@{}}#2\end{tabular}}
\centering
\caption{Comparison between the proposed methods for a given compression target size (in bytes) on the 32x32 image benchmark.}
\label{tab:results32}
\begin{tabular}{lccc}
  \toprule
  & Patch Size & \specialcellc{SSIM / 64B Target\\ (Header-less Size)} & \specialcellc{SSIM / 128B Target\\ (Header-less Size)}\\
  \midrule
  Header-less JPEG      & 8$\times$8 & \specialcellc[c]{0.70\\(72.5 bytes avg.)} & \specialcellc[c]{0.80\\(133 bytes avg.)}\\
  Header-less JPEG 2000 &            & \specialcellc[c]{0.66\\(73 bytes avg.)}   & \specialcellc[c]{0.77\\(156 bytes avg.)}\\
  Header-less WebP      &            & \specialcellc[c]{0.62\\(80.7 bytes avg.)}   & \specialcellc[c]{0.73\\(128.2 bytes avg.)}\\
  \midrule
  \specialcelll{Fully Connected Residual Encoder\\ (Shared Weights)}   & 8$\times$8            & 0.46          & 0.48          \\
  \specialcelll{Fully Connected Residual Encoder\\ (Distinct Weights)} & 8$\times$8            & 0.65          & 0.75          \\
  \textbf{LSTM Compressor}                                            & \textbf{8$\times$8}   & \textbf{0.69} & \textbf{0.81} \\
  \midrule
  \specialcelll{Conv/Deconv Residual Encoder\\ (Shared Weights)}       & 32$\times$32          & 0.45          & 0.46          \\
  \specialcelll{Conv/Deconv Residual Encoder\\ (Distinct Weights)}     & 32$\times$32          & 0.65          & 0.75          \\
  Convolutional/Deconvolutional Autoencoder & 32$\times$32 & 0.76 & 0.86 \\
  \textbf{Conv/Deconv LSTM Compressor}                                & \textbf{32$\times$32} & \textbf{0.77} & \textbf{0.87} \\
  \bottomrule
\end{tabular}
\end{table*}
\endgroup

When analyzing the results, a higher score implies a better reconstruction,
with a score of 1.0 representing a perfect reconstruction. The lowest possible
score is 0.0.
Note that while there are other metrics \citep[e.g.,][]{psnrhvsm} which emulate
the human visual system better than SSIM, we chose to use SSIM here due to its
ubiquity and ease of comparison with previous work.

\subsection{32\texorpdfstring{$\times$}{x}32 Benchmark}

Our 32$\times$32 benchmark dataset contains 216 million random color images
collected from the public internet. To be included in the dataset, each image
must originally have more than 32 pixels on both axes.
Qualified images were then downsampled to 32$\times$32, losing their original aspect ratios.
This downsampling eliminates pre-existing compression artifacts for most images. The
final 32$\times$32 images were then stored losslessly (as PNG) before being
used for training and testing. For training the LSTM models, $90\%$ of the
images were used; the remaining $10\%$ were set aside for evaluation. For
evaluating the image codecs, we use a subset of this data containing 100k random
images.

\autoref{tab:results32} summarizes the results on the 32$\times$32 benchmark,
comparing our two LSTM approaches to two JPEG codecs and to WebP. To avoid unfairly
penalizing the codecs due to the unavoidable cost of their file headers,
we exclude the header size from all metrics. Note also that since these
standard codecs can not be tuned to an exact byte budget (e.g., 64 bytes excluding the file
header), we search for the encoder quality setting that leads to a file whose
size is as close as possible, but never less than, the target size. On
average, this leads to each JPEG and WebP image consuming slightly more space than we
allow for the LSTM models.

\subsection{Analysis}

\begin{figure}[tbp]
  \newlength\imagewidth
  \setlength\imagewidth{.18\linewidth}
  \centering
  \begin{subfigure}[t]{.19\linewidth}
    \centering
    \includegraphics[width=\imagewidth]{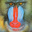}
    \caption*{Original (32$\times$32)
    \label{fig:mandrill-original}}
  \end{subfigure}
  \hfill
  \begin{subfigure}[t]{.76\linewidth}
    \centering
    \includegraphics[width=\imagewidth]{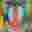}
    \includegraphics[width=\imagewidth]{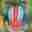}
    \includegraphics[width=\imagewidth]{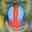}
    \includegraphics[width=\imagewidth]{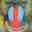}
    \caption*{JPEG compressed images
    \label{fig:mandrill-jpeg}}
  \end{subfigure}
  \\
  \begin{minipage}{.19\linewidth}
  \end{minipage}
  \hfill
  \begin{subfigure}[t]{.76\linewidth}
    \centering
    \includegraphics[width=\imagewidth]{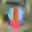}
    \includegraphics[width=\imagewidth]{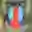}
    \includegraphics[width=\imagewidth]{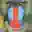}
    \includegraphics[width=\imagewidth]{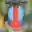}
    \caption*{WebP compressed images
    \label{fig:mandrill-webp}}
  \end{subfigure}
  \\
  \begin{minipage}{.19\linewidth}
  \end{minipage}
  \hfill
  \begin{subfigure}[t]{.76\linewidth}
    \centering
    \includegraphics[width=\imagewidth]{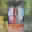}
    \includegraphics[width=\imagewidth]{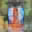}
    \includegraphics[width=\imagewidth]{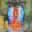}
    \includegraphics[width=\imagewidth]{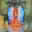}
    \caption*{Compressed images with LSTM architecture
    \label{fig:mandrill-woody}}
  \end{subfigure}
  \\
  \begin{minipage}{.19\linewidth}
  \end{minipage}
  \hfill
  \begin{subfigure}[t]{.76\linewidth}
    \centering
    \includegraphics[width=\imagewidth]{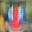}
    \includegraphics[width=\imagewidth]{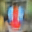}
    \includegraphics[width=\imagewidth]{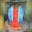}
    \includegraphics[width=\imagewidth]{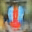}
    \caption*{Compressed images with conv/deconv LSTM architecture
    \label{fig:mandrill-dd}}
  \end{subfigure}
  \\[2ex]
  \begin{tabular}{llllll}
    \toprule
    & & \multicolumn{4}{c}{From left to right}\\
    \midrule
    %% \multirow{3}{*}{Average bits per pixel (bpp)} & JPEG & 1.050 & 1.150 & 1.270 & 1.380 \\
    %%                                               & WebP & 1.008 & 1.148 & 1.242 & 1.398 \\
    %%                                               & LSTM & 1.000 & 1.125 & 1.250 & 1.375 \\
    %%                                               & (De)Convolutional LSTM & 1.000 & 1.125 & 1.250 & 1.375 \\
    \multirow{3}{*}{Average bits per pixel (bpp)} & JPEG & 0.641 & 0.875 & 1.117 & 1.375 \\
                                                  & WebP & 0.789 & 0.914 & 1.148 & 1.398 \\
                                                  & LSTM & 0.625 & 0.875 & 1.125 & 1.375 \\
                                                  & (De)Convolutional LSTM & 0.625 & 0.875 & 1.125 & 1.375 \\
    \bottomrule
  \end{tabular}
  \caption{32$\times$32 image compression comparison between JPEG and convolutional/deconvolutional LSTM architecture.}
  \label{fig:mandrill}

\end{figure}

These 32$\times$32 images contain considerable detail that is perceptually
relevant. As can be seen in \autoref{fig:mandrill}, compressing these images
without destroying salient visual information or hallucinating false details
is challenging. At these very low bitrates and spatial resolution, JPEG block
artifacts become extremely prominent, and WebP either introduces blocking or
overly blurs the image depending on the strength of the internal filter. Color
smearing artifacts due to the codecs' default (4:2:0) chroma subsampling are
also clearly visible.

\begin{figure*}[tbp]
  \centering
  \includegraphics[width=0.98\linewidth]{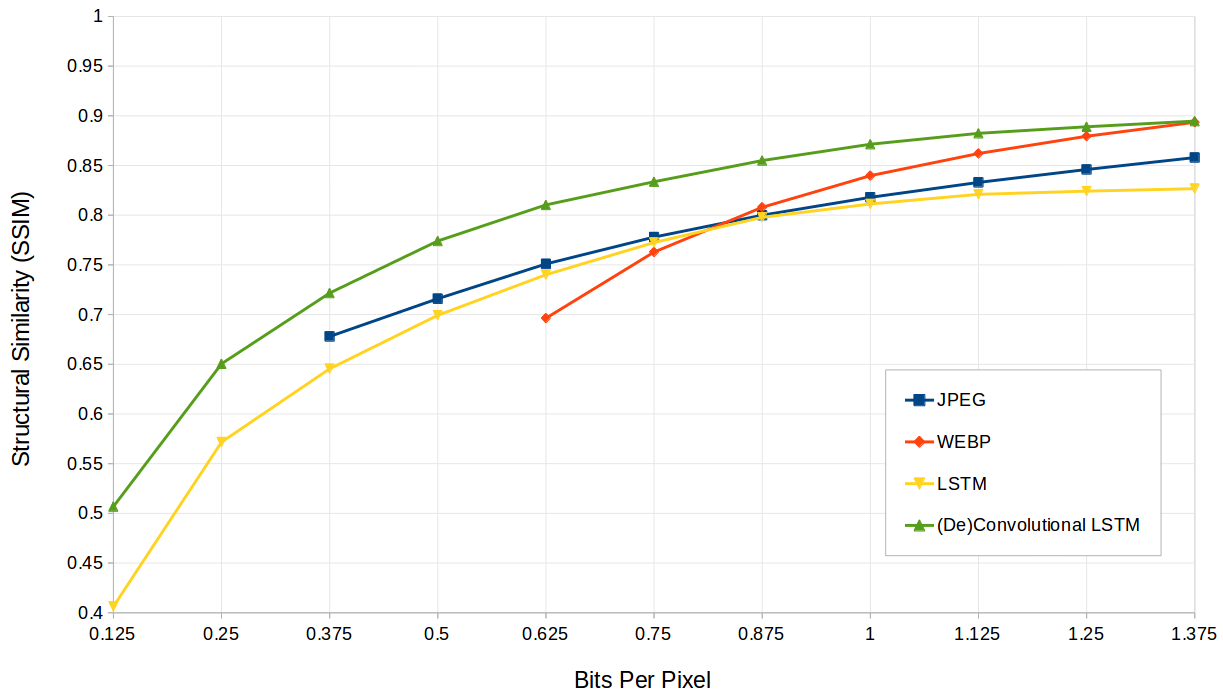}
  \caption{Rate-distortion graph showing the SSIM for different codecs at
    different target bit rates. The results are averaged over 100k images for
    JPEG and WebP and over a 10\% hold-out set of over 21 million images for
    the two LSTM models. The (de)convolutional LSTM model provides the highest
    SSIM across all low bit rates, though we expect WebP to provide better
    SSIM at higher bit rates.}
  \label{fig:rate-distortion}
\end{figure*}

Compared to JPEG, the non-convolutional LSTM model slightly reduces
inter-block boundaries on some images but can also lead to increased color
bleeding (e.g., on mandrill as shown in \autoref{fig:mandrill}). Furthermore,
the visual quality never exceeds JPEG on average as measured by SSIM and shown
in \autoref{fig:rate-distortion}. This motivates the (de)convolutional LSTM
model, which eliminates block artifacts while avoiding excessive smoothing. It
strikes the best balance between preserving real detail and avoiding color
smearing, false gradients, and hallucinated detail not present in the original
image.

Note that the (de)convolutional LSTM model exhibits perceptual quality levels
that are equal to or better than both JPEG and WebP at $4\%$ -- $12\%$ lower
average bitrate. We see this improvement despite the fact that, unlike JPEG
and WebP, the LSTMs do not perform chroma subsampling as a
preprocess. However, at the JPEG quality levels used in
\autoref{fig:mandrill}, disabling subsampling (i.e., using 4:4:4 encoding)
leads to a costly increase in JPEG's bitrate: $1.32$-$1.77$~bpp instead of
$1.05$-$1.406$~bpp, or $26\%$ greater.  This means that if we desired to
preserve chroma fidelity, we would need to drastically reduce JPEG encoding
quality in order to produce 4:4:4 JPEGs at a comparable bitrate to the LSTM
models.

In terms of coding efficiency, we took an autoencoder architecture (one
iteration of the model presented in \autoref{sub:dewey}) with a given bit
budget of either 64 or 128 bytes, and compared its SSIM against the
(de)convolutional LSTM encoder at these targets. In both cases, the LSTM model
produces SSIM values that are equivalent to the autoencoder, even though the
resulting model is more flexible.

% Rate-Distortion Graph

\vspace{-0.5em}
\section{Conclusion \& Future Work}
\vspace{-0.5em}

We describe various methods for variable-length encoding of image patches using
neural networks, and demonstrate that for the given benchmark, the fully-connected
LSTM model can perform on par with JPEG, while the convolutional/deconvolutional
LSTM model is able to significantly outperform JPEG on the SSIM perceptual metric.

While our current approach gives favorable results versus modern codecs on small
images, codecs that include an entropy coder element tend to improve (in a
bits-per-pixel sense) with greater resolution, meaning that by choosing an
arbitrarily large test image it is always possible to defeat an approach like
that described in this work. Therefore, an obvious need is to extend the current
work to function on arbitrarily large images, taking advantage of spatial
redundancy in images in a manner similar to entropy coding.

Although we presented a solution for dynamic bit assignment in the convolutional
case, it is not a fully satisfactory solution as it has the potential to introduce
encoding artifacts at patch boundaries. Another topic for future work is
determining a dynamic bit assignment algorithm that is compatible with the
convolutional methods we present, while not creating such artifacts.

The algorithms that we present may also be extended to work on video, which
we believe to be the next grand challenge for neural network-based compression.

\bibliography{biblio}
\bibliographystyle{iclr2016_conference}
\newpage
\section{Appendix: Bitwise Encoding \& Decoding}
\begin{figure}[tbp]
  \setlength\imagewidth{.4\linewidth}
  \centering
  \begin{subfigure}[t]{.95\linewidth}
    \centering
    \includegraphics[width=\imagewidth,clip,trim={5cm 5cm 5cm 5cm}]{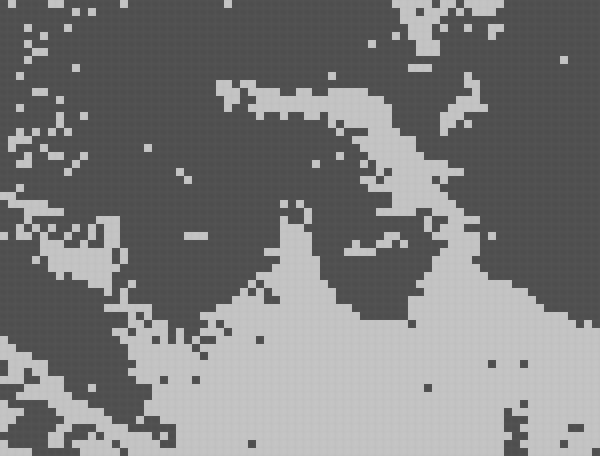}
    \includegraphics[width=\imagewidth,clip,trim={5cm 5cm 5cm 5cm}]{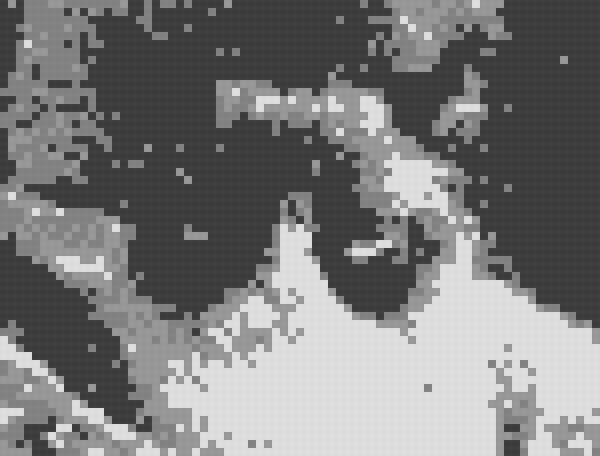}
    \includegraphics[width=\imagewidth,clip,trim={5cm 5cm 5cm 5cm}]{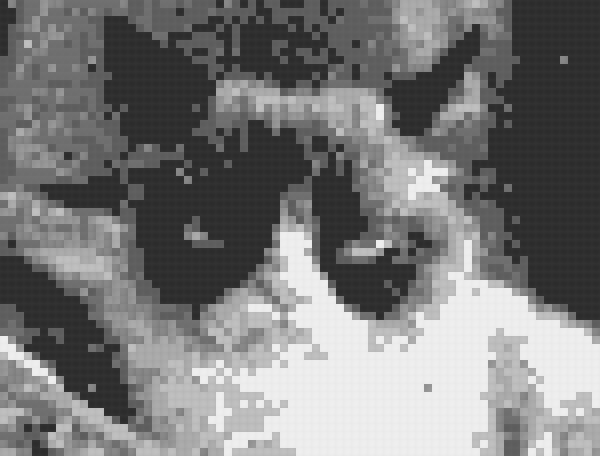}
    \includegraphics[width=\imagewidth,clip,trim={5cm 5cm 5cm 5cm}]{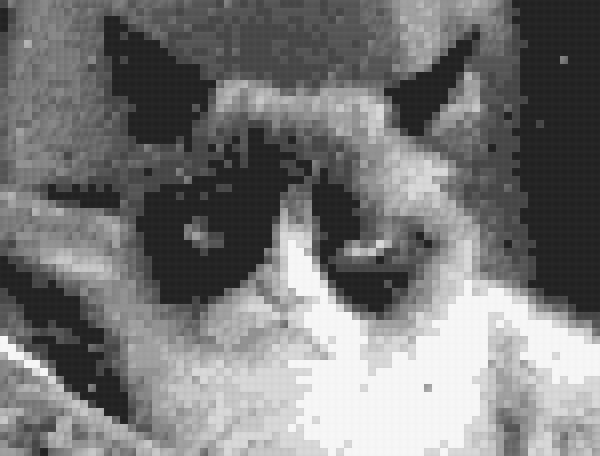}
  \end{subfigure}
  \\[2ex]
  \caption{The effect of the first four bits on compressing a cat image.
  The image on the top left has been created by using a single bit for each 8$\times$8
  block. The subsequent images add one additional bit to be processed
  by the LSTM decoder (the ordering is top-left going to bottom-right).
  The final image (bottom right) has been created by running four steps
  of the algorithm, thus allowing a total of four bits to be used to encode
  each 8$\times$8 block.}
  \label{fig:grumpy}
\end{figure}

\begin{figure}[t]
  \setlength\imagewidth{.18\linewidth}
  \centering
  \begin{subfigure}[t]{.75\linewidth}
    \centering
    \includegraphics[width=\imagewidth]{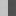}
    \includegraphics[width=\imagewidth]{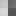}
    \includegraphics[width=\imagewidth]{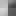}
    \includegraphics[width=\imagewidth]{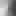}
  \end{subfigure}
  \\[2ex]
  \caption{Four 8$\times$8 blocks are encoded one bit at a time using the fully connected LSTM model.
  The blocks were encoded with 1, 4, 8, 12 bits (from left to right).}
  \label{fig:bits}
\end{figure}

In order to better understand the network architecture proposed
in~\autoref{sub:woody}, we initially limited it in terms of its capacity (bottleneck size)
and target (complexity of reconstruction). Namely, we restricted the output per step to one bit, and trained the network to
compress grayscale images. We took this simpler network and encoded a
popular image of a cat one bit at a time. \autoref{fig:grumpy} shows the effect
of the first four steps of this encoding.

\autoref{fig:bits} depicts the behavior of additional bits on four 8$\times$8
blocks from the cat image using the same network. In this zoomed-in version
it is apparent that the network first learns to differentiate between
\enquote{dark} and \enquote{light} patches using the first bit. Given an additional bit,
the network is able to introduce new solid shades of gray. One more bit starts introducing
simple gradients, which are further refined with a fourth bit, and so on.
% It is interesting to note that given this behavior it is possible to
% produce negative images simply by inverting the bit stream.

\end{document}